\documentclass[letterpaper, 10 pt, conference]{ieeeconf}  

\IEEEoverridecommandlockouts                              

\usepackage[utf8]{inputenc} 
\usepackage[T1]{fontenc}    
\usepackage{hyperref}       
\usepackage{nicefrac}       
\usepackage{microtype}      
\usepackage{xcolor}         
\usepackage{multicol}
\usepackage{multirow}
\usepackage{times}
\usepackage{soul}
\usepackage{url}
\usepackage[small]{caption}
\usepackage{graphicx}
\usepackage{amsmath}

\usepackage{booktabs}
\usepackage{algorithm}
\usepackage{algorithmic}
\usepackage{amsmath, amssymb}
\usepackage{subcaption}
\usepackage{graphicx}
\usepackage{wrapfig} 
\usepackage{tabularx}
\usepackage{textcomp} 
\usepackage{marvosym}
\title{Frequency Domain Enhanced U-Net for Low-Frequency Information-Rich Image Segmentation in Surgical and Deep-Sea Exploration Robots}

\author{Guohao Huo$^{1}$, Ruiting Dai$^{1}$, Jinliang Liu$^{1}$, Ling Shao$^{2}$, Hao Tang$^{3}$\textsuperscript{,\Letter}%
\thanks{\textsuperscript{\Letter}Corresponding author}%
\thanks{$^{1}$Guohao Huo, Ruiting Dai and Jinliang Liu are with the University of Electronic Science and Technology of China. 
        {\tt\small \{gh.huo513, rtdai, lucaliu510\}@gmail.com}}%
\thanks{$^{2}$Ling Shao is with the University of Chinese Academy of Sciences. 
        {\tt\small ling.shao@ieee.org}}%
\thanks{$^{3}$Hao Tang is with School of Computer Science, Peking University. 
        {\tt\small hao.tang@vision.ee.ethz.ch}}%
}

\begin{document}

\maketitle

\begin{abstract}
In deep-sea exploration and surgical robotics scenarios, environmental lighting and device resolution limitations often cause high-frequency feature attenuation. Addressing the differences in frequency band sensitivity between CNNs and the human visual system (mid-frequency sensitivity with low-frequency sensitivity surpassing high-frequency), we experimentally quantified the CNN contrast sensitivity function and proposed a wavelet adaptive spectrum fusion (WASF) method inspired by biological vision mechanisms to balance cross-frequency image features. Furthermore, we designed a perception frequency block (PFB) that integrates WASF to enhance frequency-domain feature extraction. Based on this, we developed the FE-UNet model, which employs a SAM2 backbone network and incorporates fine-tuned Hiera-Large modules to ensure segmentation accuracy while improving generalization capability. Experiments demonstrate that FE-UNet achieves state-of-the-art performance in cross-domain tasks such as marine organism segmentation and polyp segmentation, showcasing robust adaptability and significant application potential. The code will be released soon.
\end{abstract}

\section{Introduction}
Image segmentation serves as a foundational task in computer vision, providing critical support for autonomous robotic \cite{Autonomous_robot} systems such as surgical robots and deep-sea exploration robots to operate in complex environments. By isolating key features and structural details in images, this technology has demonstrated remarkable potential in cross-domain applications such as marine organism segmentation and polyp segmentation, significantly advancing the development of surgical robotics and deep-sea exploration systems. Although specialized architectures have achieved outstanding performance, critical challenges remain in overcoming low-frequency environments caused by seawater optical attenuation \cite{Seawater} and high-frequency attenuation from surgical electrocoagulation particulates \cite{ele_fog}, both representing urgent technical bottlenecks.

Deep convolutional neural networks (CNNs) have markedly improved segmentation accuracy, yet their inherent bias toward high-frequency feature learning often leads to suboptimal performance in low-frequency dominant scenarios \cite{Frequency_related_3,Frequency_related_4}. For instance, in marine organism segmentation, underwater non-uniform illumination combined with seawater scattering/absorption/refraction induces high-frequency attenuation \cite{Seawater}. Meanwhile, in polyp segmentation, uneven endoscopic lighting and scattering effects from electrocautery smoke amplify low-frequency components while suppressing high-frequency details \cite{ele_fog}, creating precision challenges for robotic surgical systems.

\begin{figure} 
  \centering
  \includegraphics[width=0.5\textwidth]{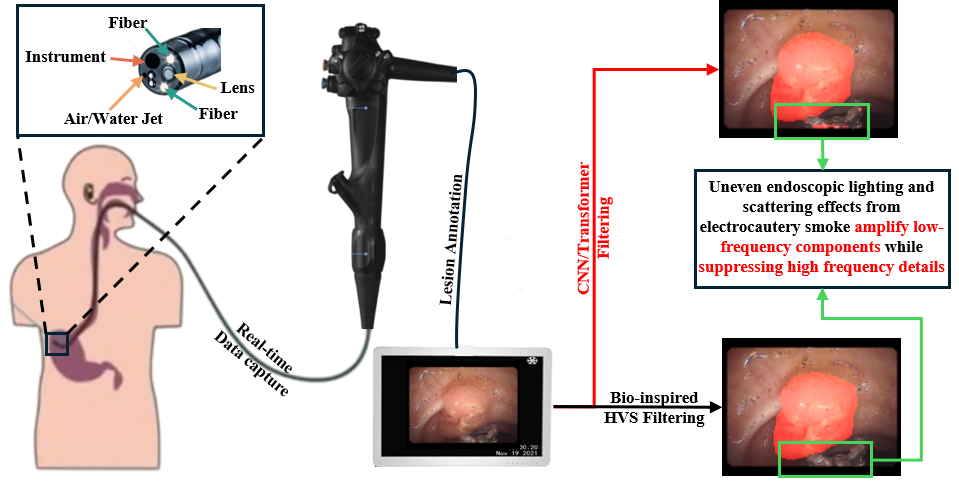}
  \caption{Under uneven illumination and electrocautery smoke–
  induced low/high-frequency distortions, compare traditional CNN/Transformer with PFB-equipped FE-UNet for polyp segmentation and lesion localization in the robotic polypectomy endoscopic acquisition and annotation workflow.}
  \label{fig:0}
    \vspace{-0.4cm}
\end{figure}

To address these issues, we propose FE-UNet - a novel feature learning framework specifically designed for low-frequency-dominant image segmentation. The framework enhances low-frequency information through Deep Wavelet Convolution (DWTConv) and employs spectral pooling filters to balance frequency components, mimicking the human visual system's mid-frequency sensitivity. The integrated Perception Frequency Block (PFB) with Wavelet-Adaptive Spectral Fusion (WASF) further enables multi-scale feature capture while simulating the receptive field-eccentricity relationship in biological vision. This synergy between CNN capabilities and visual system principles equips deep-sea exploration robots and surgical robots with human-like frequency-aware perception, significantly improving segmentation accuracy under challenging conditions.

During robotic polypectomy procedures, electrocautery-generated smoke scattering and irregular illumination from light guides cause attenuation of high-frequency details and amplification of low-frequency components in the images presented to the display, which severely impairs the ability of conventional CNN/Transformer-based segmentation methods to precisely localize lesions in the input (top right), as shown in Figure \ref{fig:0}. Our proposed FE-UNet addresses this by incorporating a bio-inspired Perception Frequency Block (PFB) for frequency-domain enhancement: the WASF module simultaneously strengthens low-frequency cues and balances high- and low-frequency features, and the enhanced representations are then fed to the decoder to produce refined segmentation masks (bottom right). The resulting masks can be fed back to the surgical robot for real-time decision-making such as localization, resection, and sampling.


In summary, our contributions are as follows:
(1) We propose FE-UNet, a frequency domain-enhanced segmentation framework designed to improve segmentation performance by enhancing low-frequency feature information in low-quality images and balancing it with high-frequency features.  
(2) We introduce PFB (Perception Frequency Block), a module that aggregates frequency-domain feature balancing mechanisms, multi-scale receptive fields, and eccentricity-aware features inspired by the human visual system, enabling autonomous robots to acquire human-like visual feature capture capabilities.
(3) We develop WASF (Wavelet-Adaptive Spectral Fusion), which strengthens low-frequency information and balances it with high-frequency features, providing a robust foundation for frequency-domain-aware feature learning.
(4) Extensive experiments on four marine animal segmentation datasets and five polyp segmentation datasets demonstrate FE-UNet's state-of-the-art performance, validating its effectiveness in addressing precise segmentation challenges in environments with dominant low-frequency information or attenuated high-frequency details.

\section{Related Work}

\subsection{Marine Animal Segmentation}
Image segmentation technology is crucial for deep-sea exploration robots in marine animal recognition tasks \cite{marine_robot}, as it enables precise identification in complex underwater environments. In recent years, convolutional neural networks (CNNs) have been widely applied in marine animal segmentation. For instance, \cite{ECDNet} proposed an Enhanced Cascaded Decoder Network (ECDNet), while \cite{Marine_Related_2} introduced a feature interaction encoder with cascaded decoders to extract more comprehensive features for accurate segmentation in challenging underwater conditions. Similarly, \cite{MASNet} designed a fusion network to learn the semantic features of marine animals.

The Segment Anything Model (SAM) has shown strong segmentation ability. Building on SAM, \cite{Dual-SAM} proposed a dual-SAM with automatic prompting to incorporate rich priors for underwater segmentation, \cite{MAS-SAM} used the SAM encoder for multi-scale features and a progressive prediction framework to capture global underwater information, and \cite{SAM2-UNet} applied a U-shaped segmentation framework to improve SAM2 for high-quality marine animal delineation.

However, CNN-based models remain limited by underwater refraction and scattering, which cause loss of high-frequency image features. Because CNNs are sensitive to high-frequency information, their ability to capture semantics is constrained under such frequency-domain distortions from light scattering and absorption.

\subsection{Polyp Segmentation}

Polyp segmentation in computer vision focuses on identifying and isolating polyp regions in medical images, which is crucial for surgical robots to recognize and differentiate polyp lesions from other tissues \cite{polyp_robot}. The main challenges arise from the diversity of polyp shapes, the ambiguity of their boundaries, and the high similarity between polyps and surrounding tissues. Reference \cite{CFANet} proposed a cross-level feature aggregation network that fuses multi-scale semantic information from different levels to achieve precise segmentation. However, this approach relies solely on convolutional neural networks (CNNs), limiting its ability to capture long-range dependencies within images.

To mitigate this, \cite{H2Former} combined CNNs and Transformers for medical segmentation. However, during surgery low-frequency interference from uneven illumination and electrocautery smoke complicates accurate operation \cite{electrocautery_smoke}, and traditional CNN-/Transformer-based models struggle to capture low-frequency edge and detail information. We propose FE-UNet, which enhances low-frequency features via wavelet transformation for refined segmentation.

\subsection{Frequency Domain Analysis}
Frequency-domain analysis has been widely applied in computer vision. Previous works \cite{Frequency_related_1,Frequency_related_2} show that low-frequency features in natural images represent global structure and color, while high-frequency features capture edges, textures, and fine details. Studies such as \cite{Frequency_related_3,Frequency_related_4} reveal that convolutional neural networks tend to bias toward high-frequency features and underrepresent low-frequency ones, whereas multi-head self-attention favors low-frequency features. WTConv \cite{Frequency_related_7} uses wavelet transforms to enhance low-frequency information and enlarge receptive fields. LITv2 \cite{Frequency_related_5} proposes the HiLo attention mixer to capture both high- and low-frequency information via self-attention, and SPAM \cite{SPANet} designs a convolutional mixer to balance high- and low-frequency signals.

\begin{figure} 
  \centering
  \includegraphics[width=0.5\textwidth]{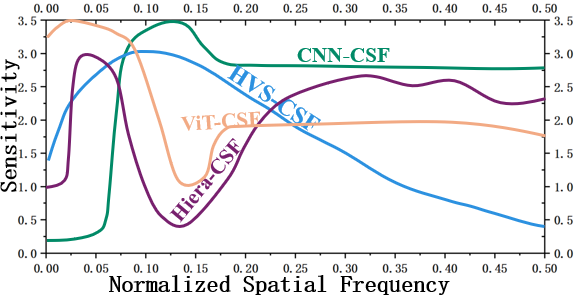}
  \caption{The Contrast Sensitivity Function model of the human visual system (HVS-CSF) and the Contrast Sensitivity Function model of convolutional neural networks (CNN-CSF), with the horizontal axis representing normalized spatial frequency and the vertical axis representing sensitivity.}
  \label{fig:1}
    \vspace{-0.4cm}
\end{figure}

To the best of our knowledge, no prior work has specifically focused on enhancing low-frequency signals while effectively balancing high- and mid-frequency information. Inspired by this, we propose a novel mixer called Wavelet-Adaptive Spectral Fusion (WASF), which utilizes Deep Wavelet Convolution (DWTConv) to enhance low-frequency signals. Subsequently, spectral pooling filters are applied to the enhanced frequency-domain features to perform frequency mixing, enabling the effective capture and utilization of high-, mid- and low-frequency information in image representations. In addition, we are the first to propose a method that simulates the human visual system based on frequency information.

\section{The Proposed Method}

\subsection{HVS-CNN Contrast Sensitivity Analysis}
The human visual system's ability to discern details is closely related to the relative contrast of the observed area, typically represented by the Contrast Sensitivity Function (CSF) \cite{CSF_3}. The CSF is a function of spatial frequency and exhibits a band-pass characteristic. Based on extensive experiments, Mannos and Sakrison proposed a classic model for the Contrast Sensitivity Function:
\begin{align}
    H(f)=2.6*(0.192+0.114)*e^{\left[-(0.114f)^{1.1}\right]},
\end{align}%
where the spatial frequency is:
\begin{align}
    f = \left( f_x^2 + f_y^2 \right)^{0.5},
\end{align}%
where $f_x$ and $f_y$ represent the spatial frequencies in the horizontal and vertical directions, respectively, based on this, we plot the contrast sensitivity function (HVS-CSF) curve of the human visual system (see Figure \ref{fig:1}).
To compare the frequency characteristics of convolutional neural networks, vision Transformers, and Hiera-L in SAM2 with those of the human visual system (HVS), we designed a simple classification experiment. The experiment uses the CIFAR-10 dataset \cite{CIFAR-10}. We performed classification inference with ResNet18, ViT-B, and Hiera-L models pretrained on ImageNet. To analyze the sensitivity of CNN, ViT, and Hiera-L feature maps to information at different frequencies, we applied the following operations sequentially to each channel of the feature maps.
\begin{equation}
\begin{aligned}
    F(u,v)&=\int_{-\infty}^\infty\int_{-\infty}^\infty f(x,y)e^{-2\pi i(ux+vy)}dxdy, \\
    M(u,v) &=
    \begin{cases}
    1 & \text{if } r \leq \mathbf{R}; \\
    0 & \text{if } r > \mathbf{R}.
    \end{cases}
\end{aligned}%
\end{equation}
Filter the image with different cutoff frequencies and then apply the inverse Fourier transform.
\begin{align}
    f_{\mathrm{filtered}}(x,y)=\int_{-\infty}^{\infty}\int_{-\infty}^{\infty}F_{\mathrm{filtered}}(u,v)e^{2\pi i(ux+vy)}dudv,
\end{align}%

Convert the frequency-domain features back to the spatial domain and then measure the model's classification accuracy at different cutoff frequencies.Plot the contrast sensitivity function curves for CNN, Transformer, and Hiera-L (CNN-CSF, ViT-CSF, and Hiera-CSF) in Figure \ref{fig:1}. We can draw the following conclusions:
\begin{enumerate}
    \item The human visual system is most sensitive to mid-frequency signals, with lower sensitivity to both low-frequency and high-frequency signals.
    \item Convolutional neural networks exhibit low sensitivity to low-frequency signals. They are more responsive to mid-to-high-frequency signals, with a slightly greater sensitivity to mid-frequency signals compared to high-frequency signals.
    \item ViT is more sensitive to low-frequency signals than to high-frequency signals, while it has relatively low sensitivity to mid-frequency signals.
    \item Hiera-L shows roughly consistent sensitivity to low- and high-frequency signals, lying between CNNs and ViT in that respect, but it has extremely low sensitivity to mid-frequency signals.
\end{enumerate}

Based on this, we propose the wavelet adaptive spectral fusion (WASF), which enhances low-frequency signals using a DWTConv. This is followed by mixing operations with a spectral pooling filter to blend high-frequency and low-frequency signals into the mid-frequency range, fully leveraging the convolutional module's high sensitivity to mid-frequency signals.

To further simulate the human visual system, we propose the Perceptual Frequency Block (PFB). This module combines frequency domain information enhancement with the relationship between receptive field and eccentricity in the human visual system, achieving a more accurate simulation of the visual perception mechanism.
Building on the PFB, the Hiera-L Block, and a U-shaped architecture, we have innovatively developed the FE-UNet architecture.

\subsection{FE-UNet}
\begin{figure*}
    \centering
    \includegraphics[width=0.865\linewidth]{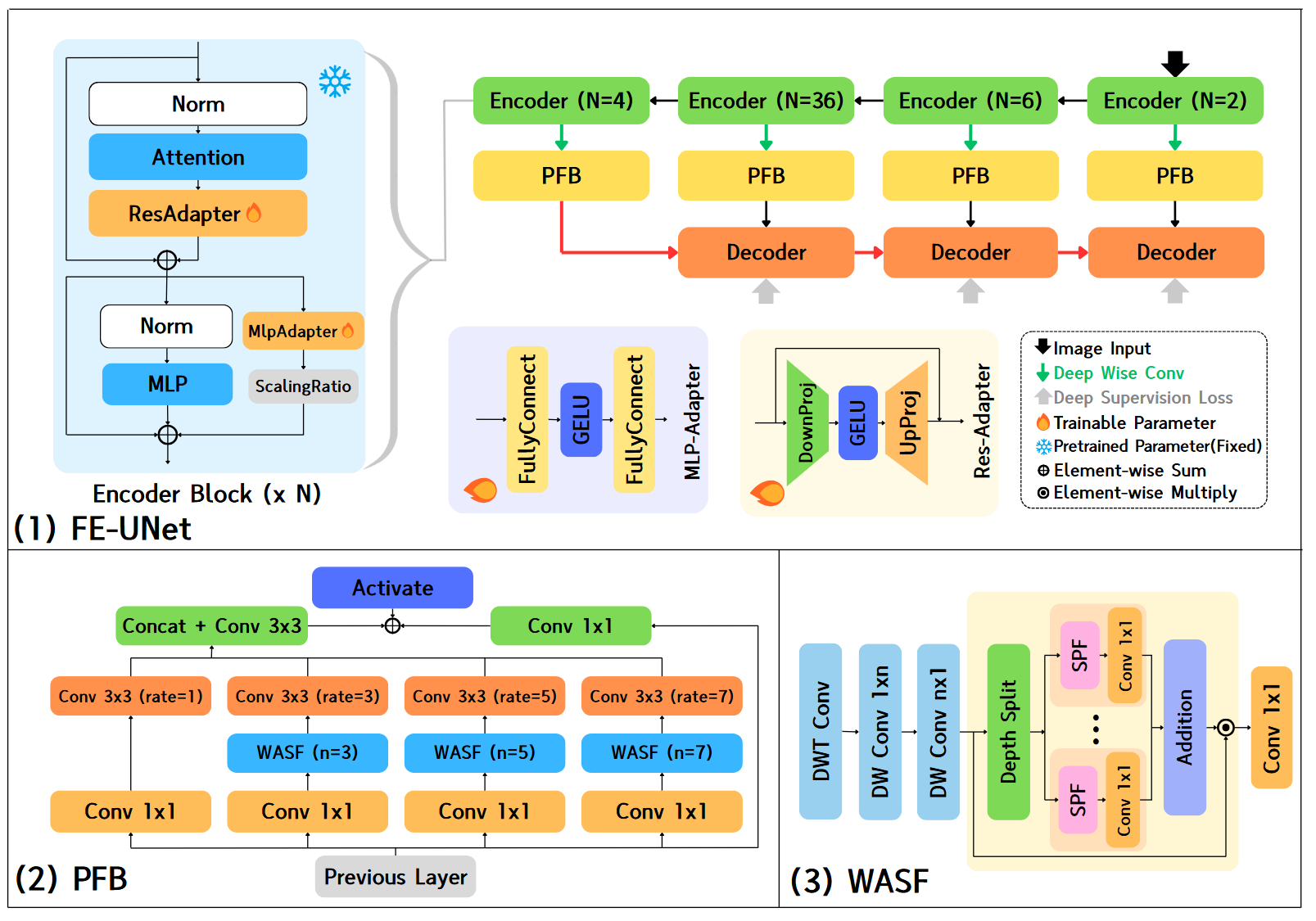}
    \caption{Fig. (a), (b), and (c) depict the architectures of the proposed FE-UNet model, Perception Frequency Block (PFB), and Wavelet-Adaptive Spectral Fusion (WASF), respectively.}
    \label{fig:2}
\end{figure*}

The original SAM2 model generates segmentation results that are class-agnostic. Without manual prompts for specific classes, SAM2 cannot produce segmentation results for designated categories. To enhance the specificity of SAM2 and better adapt it to specific downstream tasks while efficiently using pre-trained parameters, we propose the FE-UNet architecture (as shown in Figure \ref{fig:2}(a)). The architecture demonstrates robust segmentation capabilities for low-resolution images.

\noindent\textbf{Encoder.}
FE-UNet leverages the pre-trained Hiera-L backbone network from SAM2. The attention mechanisms within the Hiera backbone address the limitations of traditional convolutional neural networks in capturing long-range contextual features. Furthermore, the hierarchical structure of the Hiera module facilitates the capture of multi-scale features, making it well-suited for designing U-shaped networks.

To achieve parameter-efficient fine-tuning, we design task-specific Res-Adapter and Mlp-Adapter with trainable parameters within the Hiera Block while keeping the parameters of the Hiera Block frozen. This approach eliminates the need to fine-tune the Hiera Block, significantly reducing memory consumption.

\noindent\textbf{Res-Adapter.}
We designed the Res-Adapter following \cite{Adapter_1,Adapter_2} : a downsampling linear layer, GeLU activation, and an upsampling linear layer. Parameters are initialized to zero so the Adapter can fine-tune attention outputs while keeping changes small. We add residual connections to avoid gradient vanishing from zero initialization. This yields efficient Hiera Block fine-tuning with minimal memory overhead.

\noindent\textbf{Mlp-Adapter.}
Inspired by \cite{New_Adapter, New_Adapter2}, we propose Mlp-Adapter, a lightweight MLP module that runs in parallel with the pretrained MLP and residual connections. It preserves backbone information flow while efficiently enhancing task-specific semantic relationships and nonlinear feature interactions, improving feature refinement with minimal memory overhead.

\noindent\textbf{PFB.}
After encoder feature extraction, the U-shaped network’s hierarchical features are fused via depthwise convolution, reducing channels to 64 to lower GPU memory for the following Perceptual Frequency Block (PFB). The fused features are then processed by the PFB, which simulates characteristics of the human visual system while emphasizing feature enhancement in the frequency domain.

\noindent\textbf{Decoder.}
We made adjustments to the decoder part of the traditional UNet architecture, using the same upsampling operations. However, we implemented a customized DoubleConv module, which consists of two identical convolution—batch normalization—ReLU activation function combinations. The convolution operations use a kernel size of 3×3. Each decoder output feature is processed through a 1×1 convolutional segmentation head to generate segmentation results $S_i(i \in \left\{1, 2, 3\right\})$. These segmentation results are then upsampled and supervised against the ground truth segmentation masks.

\noindent\textbf{Loss Function.}
Each hierarchical structure loss function in FE-UNet is composed of a weighted Intersection over Union (IoU) and Binary Cross-Entropy (BCE) loss. The specific single-level loss function is defined as follows:
\begin{align}
    L = L_{IoU}^w + L_{BCE}^w.
\end{align}%
Since we employ deep supervision, the final loss function for FE-UNet is expressed as the sum of the individual hierarchical losses:
\begin{align}
    L_{total} = \sum_{i=1}^{3} L(G, S_i).
\end{align}%

\subsection{PFB}
The human visual system's ability to capture semantic information in natural images varies across different frequency domains. Generally, the human visual system (HVS) is most sensitive to mid-frequency signals, with higher sensitivity to low-frequency signals compared to high-frequency ones. 

We employ Wavelet-Adaptive Spectral Fusion (WASF) to enhance low-frequency components in image features while performing adaptive fusion operations with high-frequency information. This process shifts the frequency characteristics of image features toward the mid-frequency range, thereby achieving simulation effects that simulate the receptive fields and eccentricity mechanisms of the human visual system. It also enhances the ability of subsequent convolutional operations to extract feature information.

To achieve multi-scale receptive field capture, we employ the wavelet-adaptive spectral fusion (WASF) with different depths and convolution kernel sizes. In the WASF, $n$ represents the radius size of the low-frequency region $\mathbf{A}^{lf}$ centered at the origin, which is $2^n$, The depth convolution part of the WASF is configured with kernel sizes of 1×n and n×1. Subsequently, the padding numbers and dilation rates for the different branches of the dilated convolutions are set to $rate = {1, 3, 5,7}$. This configuration facilitates expansion of the receptive field and aligns the feature sizes, making it convenient for subsequent concatenation operations. As a result, we propose the PFB, with the structural diagram illustrated in Figure \ref{fig:2}(b).
\subsection{WASF}

In the field of computer vision, two common image filtering methods are used: one involves kernel convolution in the spatial domain, while the other utilizes the Fourier transform for filtering in the frequency domain. The method proposed in this paper operates in the frequency domain, but to achieve simple and efficient deep aggregation of spectral information under different receptive fields, we employ wavelet filtering. By applying a multi-branch spectral pooling filter followed by mixing operations on the Deep Wavelet Convolution (DWTConv), we introduce the Wavelet-Adaptive Spectral Fusion (WASF). The module architecture is shown in Figure \ref{fig:2}(c).

\noindent\textbf{DWTConv.}
To fully exploit low-frequency features, we employ cascaded deep-wavelet convolution operations. Similar to the WTConv approach (Reference \cite{Frequency_related_7}), we only adjusted the number of channels to 1.

\noindent\textbf{SPF.}
\begin{table*}[!htbp]
  \centering
  \caption{Marine animal segmentation performance on MAS3K, RMAS, UFO120 and RUWI datasets.}
  \resizebox{1\linewidth}{!}{%
  \begin{tabular}{c|r|ccccc|ccccc|ccccc|ccccc}
    \hline
    \multirow{2}{*}{Category}&\multirow{2}{*}{Method}&
    \multicolumn{5}{c|}{MAS3K}&\multicolumn{5}{c|}{RMAS}&\multicolumn{5}{c|}{UFO120}&\multicolumn{5}{c}{RUWI}\\
    \cline{3-22}
    &&mIoU&S$_{\alpha}$&F$^{w}_{\beta}$&mE$_{\phi}$&MAE&
      mIoU&S$_{\alpha}$&F$^{w}_{\beta}$&mE$_{\phi}$&MAE&
      mIoU&S$_{\alpha}$&F$^{w}_{\beta}$&mE$_{\phi}$&MAE&
      mIoU&S$_{\alpha}$&F$^{w}_{\beta}$&mE$_{\phi}$&MAE\\
    \hline
    \multirow{13}{*}{CNN}
    &PFANet \cite{PFANet}        &0.405&0.690&0.471&0.768&0.086&0.556&0.767&0.582&0.810&0.051&0.677&0.752&0.723&0.815&0.129&0.773&0.765&0.811&0.867&0.096\\
    &SCRN \cite{SCRN}        &0.693&0.839&0.730&0.869&0.041&0.695&0.842&0.731&0.878&0.030&0.678&0.783&0.760&0.839&0.106&0.830&0.847&0.883&0.925&0.059\\
    &UNet++ \cite{UNet++}        &0.506&0.726&0.552&0.790&0.083&0.558&0.763&0.644&0.835&0.046&0.412&0.459&0.433&0.451&0.409&0.586&0.714&0.678&0.790&0.145\\
    &U2Net \cite{U2-Net}         &0.654&0.812&0.711&0.851&0.047&0.676&0.830&0.762&0.904&0.029&0.680&0.792&0.709&0.811&0.134&0.841&0.873&0.861&0.786&0.074\\
    &SINet \cite{SINet}         &0.658&0.820&0.725&0.884&0.039&0.684&0.835&0.780&0.908&0.025&0.767&0.837&0.834&0.890&0.079&0.785&0.789&0.825&0.872&0.096\\
    &BASNet \cite{BASNet}        &0.677&0.826&0.724&0.862&0.046&0.707&0.847&0.771&0.907&0.032&0.710&0.809&0.793&0.865&0.097&0.841&0.871&0.895&0.922&0.056\\
    &PFNet \cite{PFNet}         &0.695&0.839&0.746&0.890&0.039&0.694&0.843&0.771&0.922&0.026&0.570&0.708&0.550&0.683&0.216&0.864&0.883&0.870&0.790&0.062\\
    &RankNet \cite{RankNet}       &0.658&0.812&0.722&0.867&0.043&0.704&0.846&0.772&0.927&0.026&0.739&0.823&0.772&0.828&0.101&0.865&0.886&0.889&0.759&0.056\\
    &C2FNet \cite{C2FNet}        &0.717&0.851&0.761&0.894&0.038&0.721&0.858&0.788&0.923&0.026&0.747&0.826&0.806&0.878&0.083&0.840&0.830&0.883&0.924&0.060\\
    &ECDNet \cite{ECDNet}        &0.711&0.850&0.766&0.901&0.036&0.664&0.823&0.689&0.854&0.036&0.693&0.783&0.768&0.848&0.103&0.829&0.812&0.871&0.917&0.064\\
    &OCENet \cite{OCENet}        &0.667&0.824&0.703&0.868&0.052&0.680&0.836&0.752&0.900&0.030&0.605&0.725&0.668&0.773&0.161&0.763&0.791&0.798&0.863&0.115\\
    &ZoomNet \cite{ZoomNet}       &0.736&0.862&0.780&0.898&0.032&0.728&0.855&0.795&0.915&\underline{0.022}&0.616&0.702&0.670&0.815&0.174&0.739&0.753&0.771&0.817&0.137\\
    &MASNet \cite{MASNet}        &0.742&0.864&0.788&0.906&0.032&0.731&0.862&0.801&0.920&0.024&0.754&0.827&0.820&0.879&0.083&0.865&0.880&0.913&0.944&0.047\\
    \hline
    \multirow{3}{*}{Trans.}
    &SETR \cite{SETR}          &0.715&0.855&0.789&0.917&0.030&0.654&0.818&0.747&0.933&0.028&0.711&0.811&0.796&0.871&0.089&0.832&0.864&0.895&0.924&0.055\\
    &TransUNet \cite{TransUNet}     &0.739&0.861&0.805&0.919&0.029&0.688&0.832&0.776&0.941&0.025&0.752&0.825&0.827&0.888&0.079&0.854&0.872&0.910&0.940&0.048\\
    &H2Former \cite{H2Former}      &0.748&0.865&0.810&0.925&0.028&0.717&0.844&0.799&0.931&0.023&0.780&0.844&0.845&\underline{0.901}&0.070&0.871&0.884&0.919&0.945&0.045\\
    \hline
    \multirow{8}{*}{SAM}
    &SAM \cite{SAM}           &0.566&0.763&0.656&0.807&0.059&0.445&0.697&0.534&0.790&0.053&0.681&0.768&0.745&0.827&0.121&0.849&0.855&0.907&0.929&0.057\\
    &Med-SAM \cite{Med-SAM}       &0.739&0.861&0.811&0.922&0.031&0.678&0.832&0.778&0.920&0.027&0.774&0.842&0.839&0.899&0.072&0.877&0.885&0.921&0.942&0.045\\
    &SAM-Adapter \cite{SAM-Adapter}   &0.714&0.847&0.782&0.914&0.033&0.656&0.816&0.752&0.927&0.027&0.757&0.829&0.834&0.884&0.081&0.867&0.878&0.913&0.946&0.046\\
    &SAM-DADF \cite{SAM-DADF}      &0.742&0.866&0.806&0.925&0.028&0.686&0.833&0.780&0.926&0.024&0.768&0.841&0.836&0.893&0.073&0.881&0.889&0.925&0.940&0.044\\
    &I-MedSAM \cite{I-MedSAM}      &0.698&0.835&0.759&0.889&0.039&0.633&0.803&0.699&0.893&0.035&0.730&0.818&0.788&0.865&0.084&0.844&0.849&0.897&0.923&0.050\\
    &Dual-SAM \cite{Dual-SAM}      &\underline{0.799}&0.884&0.838&0.933&0.023&0.735&0.860&\underline{0.812}&\underline{0.944}&\underline{0.022}&\underline{0.810}&0.856&\textbf{0.864}&\textbf{0.914}&\underline{0.064}&\underline{0.904}&\underline{0.903}&\underline{0.939}&\underline{0.959}&\textbf{0.035}\\
    &MAS-SAM \cite{MAS-SAM}       &0.788&0.887&\underline{0.840}&\underline{0.938}&0.025&\underline{0.742}&\underline{0.865}&\textbf{0.819}&\textbf{0.948}&\textbf{0.021}&0.807&\underline{0.861}&\textbf{0.864}&\textbf{0.914}&\textbf{0.063}&0.902&0.894&\textbf{0.941}&\textbf{0.961}&\textbf{0.035}\\
    &SAM2-UNet \cite{SAM2-UNet}     &\underline{0.799}&\textbf{0.903}&\textbf{0.848}&\textbf{0.943}&\textbf{0.021}&0.738&\textbf{0.874}&0.810&\underline{0.944}&\underline{0.022}&\underline{0.810}&0.858&0.845&0.889&0.072&0.906&0.903&0.927&0.941&\underline{0.037}\\
    \cline{2-22}
    &FE-UNet (Ours)&\textbf{0.815}&\underline{0.900}&\textbf{0.848}&0.928&\underline{0.022}&\textbf{0.758}&\textbf{0.874}&0.811&0.938&\textbf{0.021}&\textbf{0.821}&\textbf{0.871}&\underline{0.856}&\textbf{0.914}&0.067&\textbf{0.914}&\textbf{0.912}&0.936&\underline{0.959}&\underline{0.037}\\
    \hline
  \end{tabular}}
  \label{tab:MAS_all}
\end{table*}
Based on the inverse power law, the most important visual information in natural images is concentrated in the mid-frequency region. After using the DWTConv, we employ spectral pooling filters to perform mixing operations on the low-frequency and high-frequency components in the spectrum, thereby increasing the weight of the low-frequency components.
First, we use a two-dimensional DFT to map the features obtained after deep convolution from the spatial domain to the frequency domain:
\begin{align}
    Z = \mathcal{F}(z) \in \mathbb{C}^{H \times W}.
\end{align}%
In the above formula, $\mathcal{F}(\cdot)$ represents the two-dimensional DFT operation. Next, we perform a shifting operation to move the low-frequency components to the center of the spectrum. We then use a Fourier transform centering function to remove the remaining parts outside of the low-frequency subset.
\begin{align}
    S^{lf} =
    \begin{cases}
    \mathcal{G}(Z)(u,v), & \text{if } (u,v) \in \mathbf{A}^{lf} \\
    0, & \text{else}
\end{cases}
\end{align}%
In the above formula, $\mathcal{G}(\cdot)$is the Fourier transform centering function,$(u,v)$is a pair of positions in the frequency domain, and $\mathbf{A}^{lf}\in\mathbb{R}^{2}$ represents the low-frequency region centered at the origin.

High-pass filters are the opposite of low-pass filters, so high-frequency components can be directly obtained by removing low-frequency components from the input feature map:
\begin{align}
    S^{hf}=\mathcal{G}(Z)-S^{lf}.
\end{align}%
Finally, by sequentially applying the inverse transformation and inverse DFT operation to the high-frequency and low-frequency components, we can obtain the spectral pooled feature map:
\begin{align}
    f_{lp}(Z) &= \mathcal{F}^{-1}(\mathcal{G}^{-1}({S}^{lf})) \in \mathbb{R}^{H \times W}, \\
    f_{hp}(Z) &= \mathcal{F}^{-1}(\mathcal{G}^{-1}({S}^{hf})) \in \mathbb{R}^{H \times W}.
\end{align}%
We mix the visual features of different frequency bands obtained from the decomposition using a combination filter, which can be represented by the following formula:
\begin{align}
    \tilde{Z} = \lambda f_{lp}(Z) + (1 - \lambda) f_{hp}(Z) \in \mathbb{R}^{H \times W}.
\end{align}%
Since $\mathcal{F(\cdot)}$ and $\mathcal{G}(\cdot)$, as well as their inverses, are linear operations, they satisfy the principle of superposition. The above formula is equivalent to:
\begin{align}
    \tilde{Z} = \mathcal{F}^{-1} \left( \mathcal{G}^{-1} \left( \lambda S^{lf} + (1 - \lambda) S^{hf} \right) \right),
\end{align}%
where $\lambda \in  [0, 1]$ is a balancing parameter. We can now manipulate the frequency spectrum of visual features by adjusting $\lambda$ to control the balance between the high-frequency and low-frequency components.

\section{Experiments}

\subsection{Datasets and Evaluation Metrics.}
Following the convention \cite{DataSet_1,SPANet}, we experimentally validated the effectiveness of FE-UNet on two tasks: marine animal segmentation and polyp segmentation. 

\textbf{Marine Animal Segmentation}
Deep-sea exploration robots are tasked with separating marine animals from their deep-sea environments to analyze biodiversity and protect endangered species. However, during operations, high-frequency attenuation caused by the combined effects of underwater non-uniform illumination and seawater scattering/absorption/refraction poses challenges. To evaluate our model's performance in low-frequency information-rich deep-sea environments, we employed four public benchmark datasets: MAS3K, RMAS, UFO120, and RUWI datasets.

\textbf{Polyp Segmentation}
Polyp segmentation aims to accurately delineate polyp regions in endoscopic images, supporting early colorectal cancer screening and intraoperative decisions. In robotic polypectomy, segmentation is challenged by wide variation in polyp shape and size, low contrast with surrounding mucosa, and imaging degradations—such as instrument motion, endoscope shake, uneven illumination, and electrocautery smoke—that cause blur and scatter, suppress high‑frequency details, and complicate boundary localization. To assess robustness and clinical applicability under these conditions, we validated the proposed model on five public benchmarks: Kvasir‑SEG, CVC‑ClinicDB, CVC‑ColonDB, CVC‑300, and ETIS, which cover diverse polyp types, acquisition settings, and degradation modes, providing a rigorous testbed for weak‑texture, low‑contrast, and motion‑artifact scenarios.

For the marine animal segmentation task, we employed five evaluation metrics—mean Intersection over Union (mIoU), S-measure ($\mathbf{S}_{\alpha}$), weighted F-measure ($\mathbf{F}^w_{\beta}$), mean E-measure ($\mathbf{mE}_{\phi}$), and mean absolute error (MAE)—to assess model performance. In the polyp segmentation task, model effectiveness was evaluated using two key metrics: mean Dice score (mDice) and mIoU.Detailed implementation details and dataset configurations are provided in the Supplementary Material.

\subsection{Comparison with State-of-the-Arts.}
In this section, we compare our method with other approaches on four public marine animal segmentation datasets and five public polyp segmentation datasets. The quantitative and qualitative results clearly demonstrate the significant advantages of our proposed method.

\begin{table}[!htbp]
\centering
\caption{Polyp segmentation performance on Kvasir-SEG \cite{kvasir}, CVC-ClinicDB \cite{ClinicDB}, CVC-ColonDB \cite{ColonDB}, CVC-300 \cite{CVC-300}, and ETIS \cite{ETIS} datasets.}
\label{tab:polyp_segmentation}
\scriptsize  
\setlength{\tabcolsep}{0.5pt} 
\begin{tabular}{r|cc|cc|cc|cc|cc}
\hline
\multirow{2}{*}{Method} & \multicolumn{2}{c|}{Kvasir} & \multicolumn{2}{c|}{ClinicDB} & \multicolumn{2}{c|}{ColonDB} & \multicolumn{2}{c|}{CVC-300} & \multicolumn{2}{c}{ETIS} \\
\cline{2-3}\cline{4-5}\cline{6-7}\cline{8-9}\cline{10-11}
& mDice & mIoU & mDice & mIoU & mDice & mIoU & mDice & mIoU & mDice & mIoU \\
\hline
UNet \cite{U-Net} & 0.818 & 0.746 & 0.823 & 0.755 & 0.504 & 0.436 & 0.710 & 0.627 & 0.398 & 0.335 \\
SFA \cite{SFA} & 0.723 & 0.611 & 0.700 & 0.607 & 0.456 & 0.337 & 0.467 & 0.329 & 0.297 & 0.217 \\
UNet++ \cite{UNet++} & 0.821 & 0.744 & 0.794 & 0.729 & 0.482 & 0.408 & 0.707 & 0.624 & 0.401 & 0.344 \\
PraNet \cite{PraNet} & 0.898 & 0.840 & 0.899 & 0.849 & 0.709 & 0.640 & 0.871 & 0.797 & 0.628 & 0.567 \\
EU-Net \cite{EU-Net} & 0.908 & 0.854 & 0.902 & 0.846 & 0.756 & 0.681 & 0.837 & 0.765 & 0.687 & 0.609 \\
SANet \cite{SANet} & 0.904 & 0.847 & 0.916 & 0.859 & 0.752 & 0.669 & 0.888 & 0.815 & 0.750 & 0.654 \\
MSNet \cite{MSNet} & 0.905 & 0.849 & 0.918 & 0.869 & 0.751 & 0.671 & 0.865 & 0.799 & 0.723 & 0.652 \\
C2FNet \cite{C2FNet} & 0.886 & 0.831 & 0.919 & 0.872 & 0.724 & 0.650 & 0.874 & 0.801 & 0.699 & 0.624 \\
MSEG \cite{MSEG} & 0.897 & 0.839 & 0.909 & 0.864 & 0.735 & 0.666 & 0.874 & 0.804 & 0.700 & 0.630 \\
DCRNet \cite{DCRNet} & 0.886 & 0.825 & 0.896 & 0.844 & 0.704 & 0.631 & 0.856 & 0.788 & 0.556 & 0.496 \\
LDNet \cite{LDNet} & 0.887 & 0.821 & 0.881 & 0.825 & 0.740 & 0.652 & 0.869 & 0.793 & 0.645 & 0.551 \\
FAPNet \cite{FAPNet} & 0.902 & 0.849 & 0.925 & 0.877 & 0.731 & 0.658 & 0.893 & 0.826 & 0.717 & 0.643 \\
ACSNet \cite{ACSNet} & 0.898 & 0.838 & 0.882 & 0.826 & 0.716 & 0.649 & 0.863 & 0.787 & 0.578 & 0.509 \\
H2Former \cite{H2Former} & 0.910 & 0.858 &0.922&0.871  & 0.719 & 0.642 & 0.856 & 0.793 & 0.614 & 0.547 \\
CaraNet \cite{CaraNet} & 0.913 & 0.859 & 0.921 & 0.876 & 0.775 & 0.700 & \underline{0.902} & \underline{0.836} & 0.740 & 0.660 \\
CFA-Net \cite{CFANet} & 0.915 & 0.861 & \underline{0.933} & \underline{0.883} & 0.743 & 0.665 & 0.893 & 0.827 & 0.732 & 0.655 \\
I-MedSAM \cite{I-MedSAM} & 0.839 & 0.759 & 0.871 & 0.788 & \textbf{0.885} & \textbf{0.800} & 0.900 & 0.822 & \textbf{0.874} & \textbf{0.791} \\
SAM2-UNet \cite{SAM2-UNet} & \underline{0.928} & \underline{0.879} & 0.907 & 0.856 & 0.808 & 0.730 & 0.894 & 0.827 & 0.796 & 0.723 \\
\hline
FE-UNet (Ours) & \textbf{0.934} & \textbf{0.889} & \textbf{0.942} & \textbf{0.895} & \underline{0.830} & \underline{0.757} & \textbf{0.914} & \textbf{0.852} & \underline{0.819} & \underline{0.748}\\
\hline
\end{tabular}
\end{table}

\noindent\textbf{Quantitative Comparison} 
\begin{figure*}[!tbp]
    \centering
    \begin{subfigure}[t]{0.52\textwidth} 
        \centering
        \includegraphics[width=\textwidth]{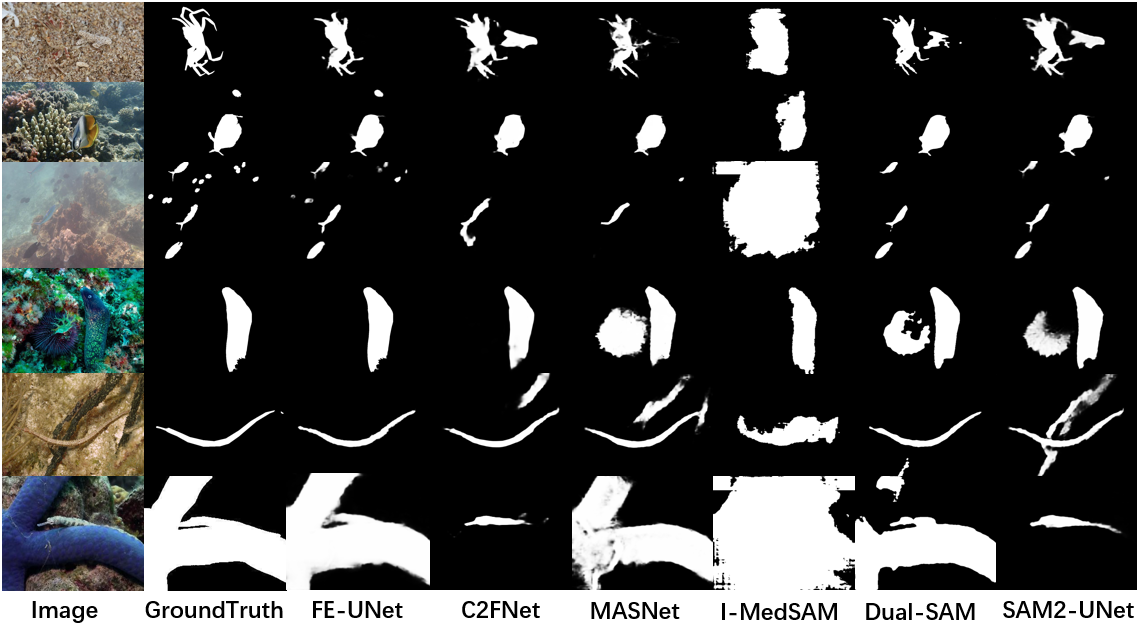}
        \caption{Marine animal segmentation task}
        \label{fig:3}
    \end{subfigure}
    \hfill 
    \begin{subfigure}[t]{0.45\textwidth} 
        \centering
        \includegraphics[width=\textwidth]{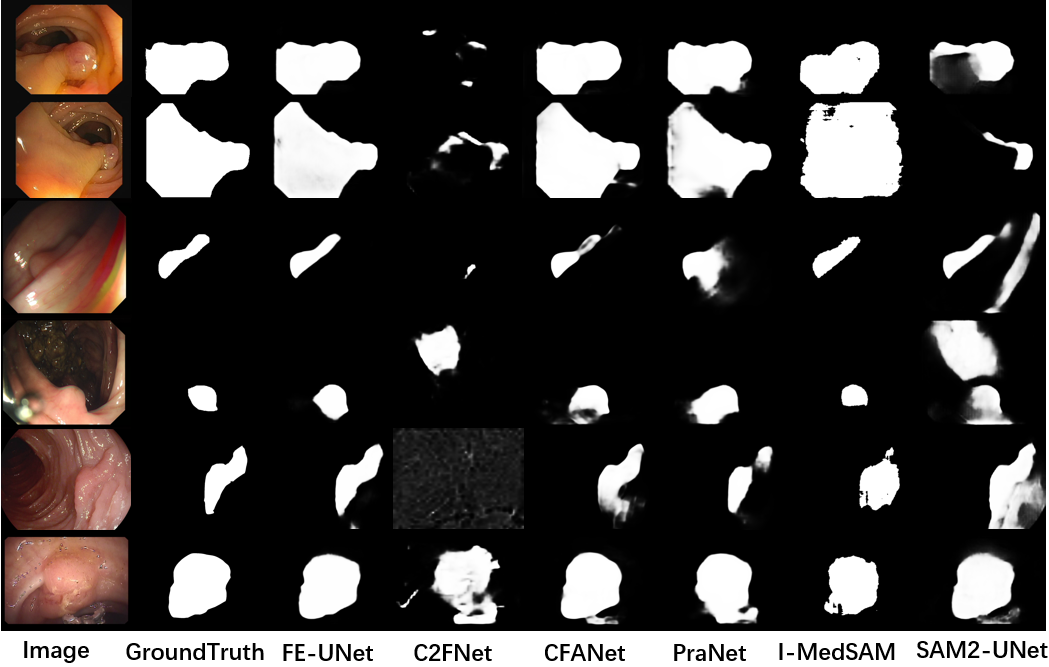}
        \caption{Polyp segmentation task}
        \label{fig:4}
    \end{subfigure}
    \caption{Qualitative results.}
    \label{fig:vis}
\end{figure*}
Tables \ref{tab:MAS_all} present quantitative comparisons of typical marine animal segmentation datasets. Compared with CNN-based methods, our method significantly improves performance. On the challenging MAS3K dataset, our method achieves the highest scores across all metrics, delivering a 4-6\% improvement. Moreover, our method consistently outperforms others on additional MAS datasets. Compared to state-of-the-art marine animal segmentation models, our model achieves a 1-2\% improvement in mIoU and $\mathbf{S}_{\alpha}$ metrics. Compared with Transformer-based methods, our method achieves a 3-6\% improvement on the MAS3K dataset. Furthermore, compared with other SAM-based methods, our model achieves a 1–2\% improvement in mIoU scores as well as $\mathbf{S}_{\alpha}$ compared to current SOTA methods.

\begin{figure*}[!tbp]
    \centering
    \begin{subfigure}[t]{0.32\textwidth} 
        \centering
        \includegraphics[width=\textwidth]{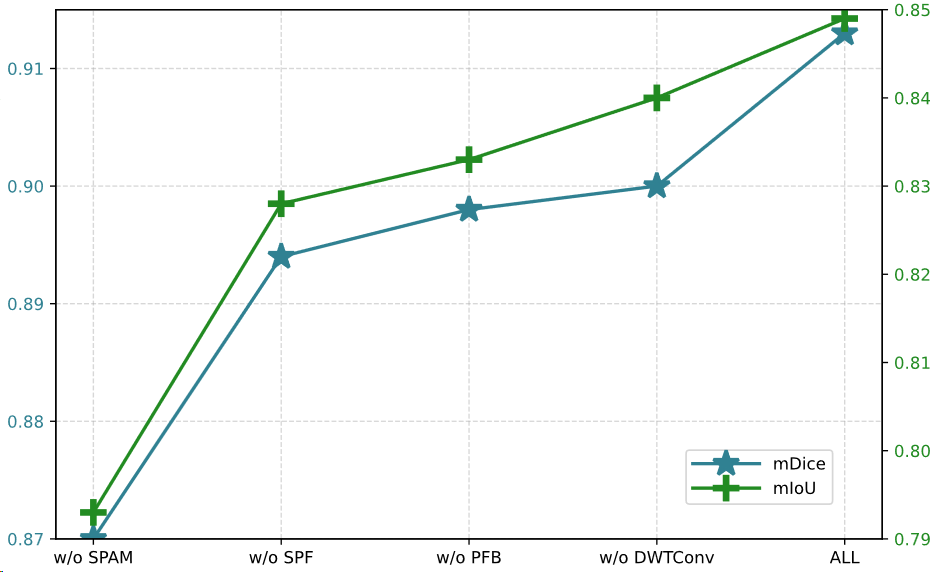}
        \caption{PFB Components}
        \label{fig:sub3}
    \end{subfigure}
    \hfill 
    \begin{subfigure}[t]{0.32\textwidth} 
        \centering
        \includegraphics[width=\textwidth]{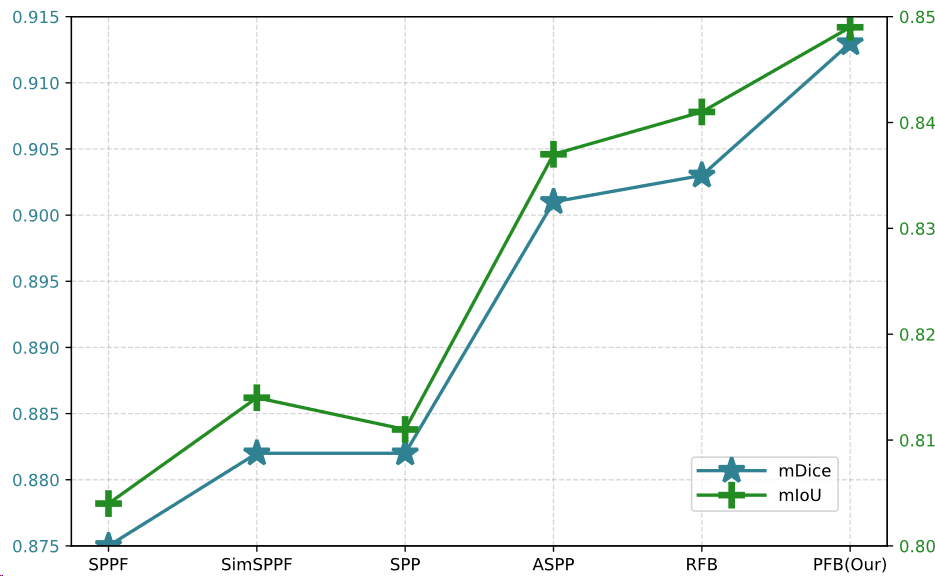}
        \caption{Other Plug-and-Play Blocks}
        \label{fig:sub4}
    \end{subfigure}
    \hfill
    \begin{subfigure}[t]{0.32\textwidth}
        \centering
        \includegraphics[width=\textwidth]{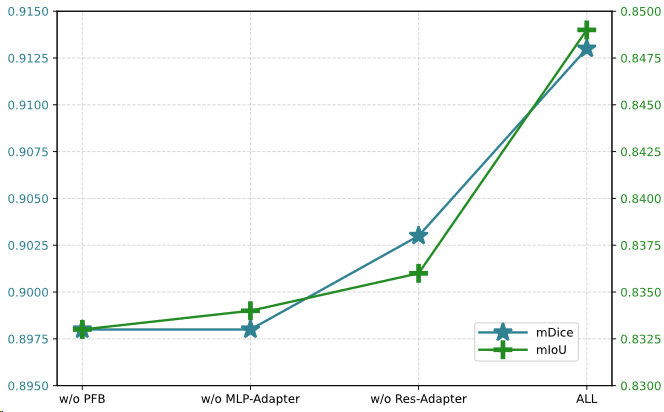}
        \caption{FE-UNet Components}
        \label{fig:sub5}
    \end{subfigure}
    \caption{Visualization of ablation experiment results.}
    \label{fig:Ablation2}
\end{figure*}

We follow \cite{SOTA_Polyp_1}, including the same comparison methods and tools. Table \ref{tab:polyp_segmentation} shows the performance of our model on five polyp segmentation test datasets. On the Kvasir, CVC-ClinicDB, and CVC-300 datasets, our model achieved SOTA performance, with a 1-2\% improvement over the second-best method. Furthermore, on the CVC-ColonDB and ETIS datasets, our model demonstrated the second-best segmentation performance.




\noindent\textbf{Qualitative Comparison} 
Figures \ref{fig:vis} illustrate some visual examples from the marine animal segmentation and polyp segmentation tasks, respectively, to further verify the effectiveness of our method. Compared with previous approaches, our method produces segmentation results that are highly similar to the ground truth in simpler tasks. Moreover, on challenging images with cluttered backgrounds and rich details, our method consistently generates more accurate and refined segmentation masks.

\subsection{Ablation Study}
As shown in Figure \ref{fig:Ablation2}, three ablation experiments on the CVC-300 dataset validated key FE-UNet components. First, ablation within the PFB module showed every element is necessary for optimal performance. Second, replacing FE-UNet’s residual multi-scale extractor with mainstream modules (SPPF \cite{yolov5}, SimSPPF \cite{yolov6}, SPP \cite{SPP}, ASPP \cite{ASPP}, RFB \cite{RFB}) demonstrated PFB’s superior multi-scale feature extraction. Third, removing PFB, MLP-Adapter, or Res-Adapter individually caused notable performance drops, confirming each component’s indispensability.

\section{Conclution}
In this work, we propose a novel feature learning framework named FE-UNet for natural image segmentation. Specifically, we introduce the Perceptual Frequency Block (PFB), which aggregates frequency-domain information enhanced by multi-scale WASF modules through the integration of multi-scale receptive fields and eccentricity-aware mechanisms. This design simulates the human visual system's heightened sensitivity to mid-frequency features, enabling our method to extract rich frequency-domain features critical for fine-grained image segmentation. Experimental results demonstrate state-of-the-art (SOTA) performance in the marine animal segmentation and polyp segmentation task. Our framework is not only applicable to marine animal and medical polyp segmentation scenarios but also lays a solid foundation for image segmentation in other complex scenarios, providing a broader research space for enhancing visual perception capabilities in autonomous robots such as surgical and deep-sea exploration systems.
\bibliography{ref}

\end{document}